\documentclass[10pt,twocolumn,letterpaper]{article}

\usepackage[pagenumbers]{cvpr} 

\definecolor{cvprblue}{rgb}{0.21,0.49,0.74}
\usepackage[pagebackref,breaklinks,colorlinks,allcolors=cvprblue]{hyperref}

\usepackage{multirow}
\usepackage{graphicx}
\usepackage{wrapfig, lipsum}
\usepackage[dvipsnames]{xcolor}
\usepackage{pifont}
\usepackage{enumitem}
\usepackage{color}
\usepackage{makecell}
\usepackage{adjustbox}

\newcommand{\cmark}{\checkmark}

\def\paperID{3053} 
\def\confName{CVPR}
\def\confYear{2025}

\newcommand{\method}{\textsc{RoboGround}\xspace}

\title{\method: Robotic Manipulation with Grounded Vision-Language Priors}

\author{
    Haifeng Huang$^{1,2}$$^*$ \hspace{0.8em} Xinyi Chen$^{2}$$^*$ \hspace{0.8em} Yilun Chen$^{2}$$^\ddagger$ \hspace{0.8em} Hao Li$^{2}$ \hspace{0.8em} Xiaoshen Han$^{2}$\\
    \hspace{0.8em} Zehan Wang$^{1}$ \hspace{0.8em} Tai Wang$^{2}$ \hspace{0.8em} Jiangmiao Pang$^{2}$ \hspace{1em} Zhou Zhao$^{1,2}$$^\dag$\\
    $^{1}$Zhejiang University \hspace{1em} $^{2}$Shanghai AI Laboratory\\
    {\tt\small \{huanghaifeng\}@zju.edu.cn}
}

\begin{document}
\maketitle

\renewcommand{\thefootnote}{}
\footnotetext{$^*$ Equal contribution.}
\footnotetext{$^\ddagger$ Project lead.}
\footnotetext{$^\dag$ Corresponding author.}
\footnotetext{\textbullet~Work done during an internship at Shanghai AI Laboratory.}


\begin{abstract}
Recent advancements in robotic manipulation have highlighted the potential of intermediate representations for improving policy generalization. In this work, we explore grounding masks as an effective intermediate representation, balancing two key advantages: (1) effective spatial guidance that specifies target objects and placement areas while also conveying information about object shape and size, and (2) broad generalization potential driven by large-scale vision-language models pretrained on diverse grounding datasets. We introduce \method, a grounding-aware robotic manipulation system that leverages grounding masks as an intermediate representation to guide policy networks in object manipulation tasks. To further explore and enhance generalization, we propose an automated pipeline for generating large-scale, simulated data with a diverse set of objects and instructions. Extensive experiments show the value of our dataset and the effectiveness of grounding masks as intermediate guidance, significantly enhancing the generalization abilities of robot policies. Code and data are available at \href{https://robo-ground.github.io}{robo-ground.github.io}.

\end{abstract}

\section{Introduction}
\label{sec:intro}

The early low-level policies for manipulation \cite{aloha,chi2023diffusion,robomimic,tian2024seer} primarily relied on imitation learning from collected demonstrations, focusing on acquiring specific skills within predefined scenes. Consequently, these policies exhibited limited generalization capabilities. More recently, Vision-Language-Action (VLA) models~\cite{RT-2, openvla} have emerged as a promising approach for learning generalizable robotic policies. These models leverage large-scale robot training data alongside pretrained Vision-Language Models (VLMs)~\cite{liu2023llava, zhu2023minigpt, glamm} to develop broad manipulation capabilities. However, it is still challenging for these methods to generalize to novel settings without access to extensive datasets and additional fine-tuning, both of which are costly.

\begin{figure}[t]
\centering
\includegraphics[width=0.45\textwidth]{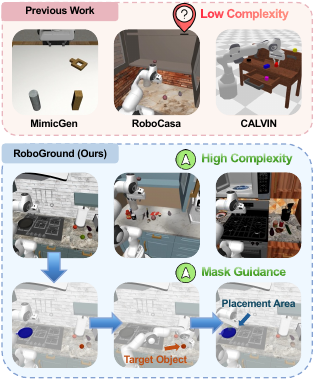}
\caption{\textbf{Examples of generated data and mask guidance for manipulation policy.} The generated data includes more object distractors in the scene, leading to higher scene complexity. The robot policy is guided by masks to localize the target object and placement area.}
\label{fig:intro}
\end{figure}

Intermediate representations \cite{rt-trajectory,rt-h,track2act,sundaresan2024rt} offer an alternative approach to generalized policy learning by providing structured guidance to policy networks across diverse scenarios. Research in this area typically falls into two categories: \textit{accessible yet coarse-grained representations}, such as language instructions~\cite{rt-h, sprint}, which are easy to generate but often lack the spatial precision required for fine-grained object manipulation; and \textit{fine-grained but resource-intensive representations}, such as goal images~\cite{black2023zero} or point flows~\cite{im2flow2act, track2act, seita2023toolflownet, eisner2022flowbot3d, yuan2024generalflow}, which provide detailed spatial guidance but demand extensive training data and computational resources, limiting their scalability.

In this work, we introduce grounding masks as a promising intermediate representation that balances two key aspects: (1) \textit{Effective spatial guidance}, which not only specifies target objects and placement areas but also conveys information about object shape and size, allowing low-level policies to accurately interpret spatial information; and (2) \textit{Broad generalization potential}, facilitated by large-scale vision-language models~\cite{glamm, lisa, florence-2, ferret-v2} pretrained on diverse grounding datasets, enhancing adaptability to novel objects and environments. To fully explore grounding mask representations for low-level policy networks, we propose \textbf{\method}, a grounding-aware system guided by object masks for robotic manipulation. This system employs a grounded vision-language model to generate masks for both the target object and placement area, which are then seamlessly integrated into the policy network through channel-level and patch-level designs. Specifically, we enhance image inputs via channel concatenation with the masks and introduce a grounded perceiver that attends to mask-guided regions at the patch level, preserving essential spatial information for precise manipulation.

As shown in Figure~\ref{fig:intro}, existing datasets~\cite{mimicgen, calvin, robocasa, gensim, optimus} often suffer from limited instruction diversity and scene complexity, leading policy networks to overfit to specific conditions and straightforward instruction-scene mappings. In this work, we aim to investigate the effectiveness of grounding mask representations in maintaining consistent performance across diverse scenarios and instructions. To address dataset limitations, we propose \textbf{an automated pipeline} for generating simulated manipulation data with a diverse set of objects and instructions. By systematically increasing scene complexity through the inclusion of distractor objects, we create a challenging dataset that promotes generalization across varied instructions and object configurations. Specifically, we generate 24K demonstrations with 112K diverse instructions, covering object appearance, spatial relationships, and commonsense knowledge. Scene complexity is further enhanced by sampling distractor objects from a large pool of 3,526 objects spanning 176 categories. This high-complexity dataset encourages models to leverage intermediate object mask guidance for interpreting instruction-scene relationships, providing a rigorous evaluation of our mask-guided model's generalization potential.

We conduct extensive experiments to evaluate the model’s generalization across diverse instructions, unseen objects and categories, and core robotic skills. Our results highlight the effectiveness of grounding masks as intermediate guidance, showing that they substantially enhance the generalization capabilities of robot policies while demonstrating the value of our dataset.


\section{Related Work}
\label{sec:related}


\noindent\textbf{Intermediate Representations for Robot Policy.}  
Various intermediate guidance strategies have been explored to enhance policy learning, including language instructions~\cite{rt-h, sprint}, 2D trajectories~\cite{rt-trajectory, wen2023atm, robotap}, point flows~\cite{im2flow2act, track2act, seita2023toolflownet, eisner2022flowbot3d, yuan2024generalflow}, goal images~\cite{black2023zero, bu2024closed}, goal sketches~\cite{sundaresan2024rt}, and object-centric representations~\cite{moo, kite, fangandliu2024moka}. For instance, RT-Trajectory~\cite{rt-trajectory} utilizes rough trajectory sketches to guide policies in performing new tasks, while Im2Flow2Act~\cite{im2flow2act} predicts object flows and maps them to actions. However, acquiring such intermediate guidance during evaluation is often impractical due to the lack of large-scale, diverse datasets and generalizable foundation models. Some approaches leverage object-centric representations to enhance policy robustness. KITE~\cite{kite} decomposes tasks into keypoints using an additional grounding module, while MOKA~\cite{fangandliu2024moka} employs prompt engineering to derive keypoints that effectively capture affordances in manipulation tasks. Closely related to our mask-based approach, MOO~\cite{moo} uses pretrained Vision-Language Models (VLMs) to generate coarse bounding boxes, which are then overlaid onto the original image. In contrast, our method focuses on obtaining fine-grained object masks and introduces an efficient Grounded Perceiver to better utilize mask-based intermediate guidance for improved manipulation performance.

\begin{figure*}[t]
    \centering
    \includegraphics[width=0.95\textwidth]{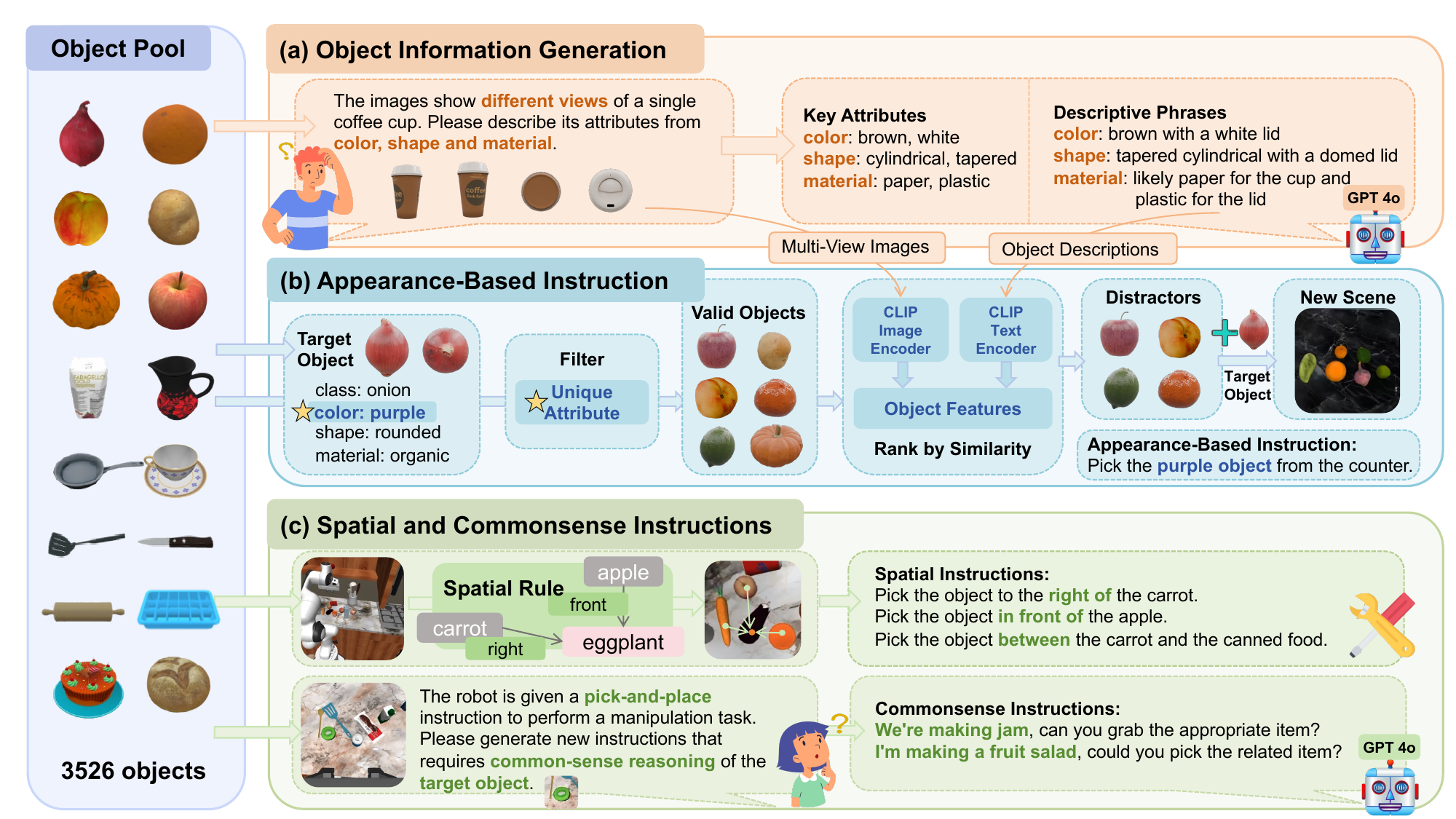}
    \caption{\textbf{Data Generation Pipeline.} The pipeline is composed of three key stages: (a) First, we extract informative object attributes in both keyword and descriptive phrase formats; (b) Next, appearance-based instructions are generated using these attributes, where keywords filter objects and descriptive phrases calculate appearance similarity; (c) Finally, spatial and commonsense instructions are generated through rule-based methods and GPT-generated techniques, respectively.}
    \label{fig:data_gen}
\end{figure*}

\noindent\textbf{Large Vision-Language Models.}  
Large Vision-Language Models (LVLMs) have recently demonstrated remarkable capabilities in image understanding and image-to-text generation, as seen in BLIP-2~\cite{li2023blip2}, LLaVA~\cite{liu2023llava}, InstructBLIP~\cite{instructblip}, and MiniGPT-4~\cite{zhu2023minigpt}. While these models excel at text generation and holistic image comprehension, they often struggle with fine-grained understanding of local regions within images. To address this, several methods~\cite{chen2023shikra, kosmos-2, wang2024allseeing_v2, you2023ferret} enhance region-level perception by incorporating bounding boxes and spatial location bins, allowing LMMs to focus on specific interactive areas. Furthermore, recent approaches~\cite{lisa, glamm, groma, ferret-v2, florence-2} have advanced grounding capabilities, enabling the identification of target objects or regions to support downstream tasks. Our method builds upon this progress by leveraging GLaMM~\cite{glamm} to generate grounding masks for both target objects and placement areas, providing structured guidance for the low-level policy network in robotic manipulation.


\section{Data Generation}  
Existing language-conditioned simulation datasets often suffer from limited object and environment diversity~\cite{cliport, rlbench, calvin} or lack a broad range of instructions and complex scenes~\cite{mimicgen, robocasa}, making them less effective for training robust robotic policies and evaluating model generalization. To address these limitations, we introduce an automated data generation pipeline built upon RoboCasa~\cite{robocasa}, designed to enhance object variety, instruction diversity, and scene complexity for more challenging manipulation tasks. Our approach produces 24K new demonstrations paired with 112K diverse instructions, covering 176 distinct object categories and a total of 3,526 unique objects.  

\subsection{Object Set}  
A diverse object set is essential for increasing scene complexity and enabling varied instruction generation. Expanding on the original RoboCasa dataset, we curated additional 1,017 high-quality tabletop manipulation objects from Objaverse~\cite{objaverse}, selected from a pool of over 760K items. The initial filtering process, supported by GPT-4~\cite{gpt-4}, leveraged object tags and descriptions to identify (1) tabletop-suitable items, (2) kitchen-related objects, and (3) exclude multi-item collections. GPT-4 further assisted in categorizing objects based on their characteristics. The final selection underwent manual review to ensure quality. By incorporating distractor objects from this extensive set, we create complex scenes that challenge models to accurately interpret instructions in diverse and dynamic environments.


\begin{figure*}[t]
    \centering
    \includegraphics[width=0.95\textwidth]{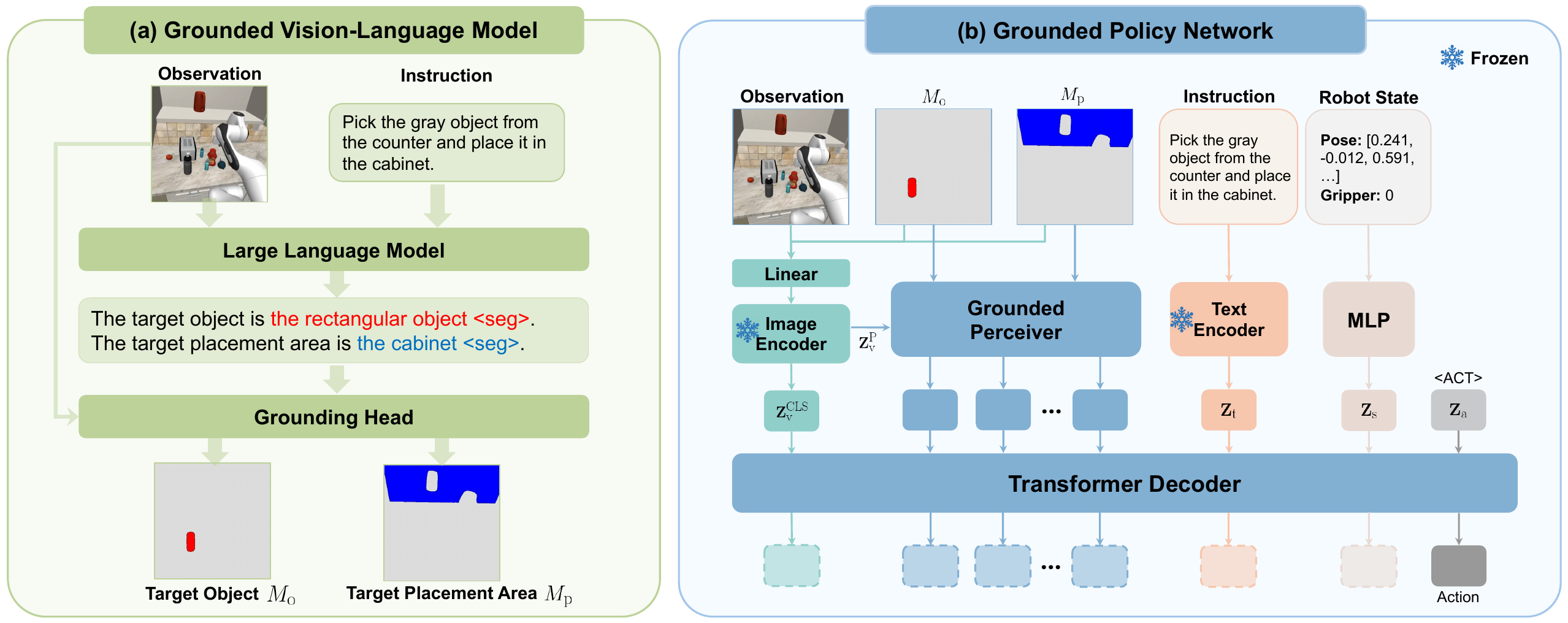}
    \caption{\textbf{Overall Architecture of \method.} To enhance policy generalization, we leverage grounding masks as intermediate representations for spatial guidance. Specifically, (a) The grounded vision-language model processes the instruction and image observation to generate target masks. (b) The grounded policy network integrates mask guidance by concatenating masks with the image input and directing attention within the grounded perceiver.}

    \label{fig:model}
\end{figure*}

\subsection{Diverse Instructions}
Existing language-conditioned datasets often rely on fixed-format instructions that require minimal reasoning, leading models to depend on simple mappings between instructions and tasks rather than developing a deeper understanding of the language's semantics. This can hinder the model's ability to generalize to novel objects or tasks. To address this limitation, we propose designing instructions that focus on discriminative understanding of object appearance, spatial relationships, and commonsense knowledge, as illustrated in Figure~\ref{fig:data_gen}. We present a target object alongside up to 10 contextually relevant objects (distractors) in the scene, encouraging the model to make precise distinctions between the target and surrounding objects.

\noindent\textbf{Appearance-Based Instructions.}
To enable detailed visual understanding, we render each object from four perspectives (front, back, left, and right) and combine these views into a composite image that offers a comprehensive visual overview. This image is then processed by GPT-4, which extracts key attributes for each object—such as class, color, shape, and material—both as keywords and full descriptive phrases. When selecting a target object, we randomly assign one attribute as its unique characteristic within the scene. Objects with overlapping keywords for this attribute are filtered out, resulting in a refined set of valid objects. From this set, distractors are sampled by ranking them based on the similarity of their textual and visual embeddings, derived from CLIP~\cite{clip} text encoder and image encoder, encoded using the full attribute descriptions and multi-view images, respectively.

\noindent\textbf{Appearance-Based Instructions.}
To enable detailed visual understanding, we render each object from four perspectives (front, back, left, and right) and combine these views into a composite image that provides a comprehensive visual overview. This image is then processed by GPT-4, which extracts key attributes for each object—such as class, color, shape, and material—both as keywords and full descriptive phrases. When selecting a target object, we randomly assign one attribute as its unique characteristic within the scene. Objects with overlapping keywords for this attribute are filtered out, resulting in a refined set of valid objects. From this set, distractors are sampled by ranking them based on the similarity of their textual and visual embeddings, derived from the CLIP~\cite{clip} text and image encoders, using the full attribute descriptions and multi-view images, respectively.

\noindent\textbf{Spatial Instructions.}
For randomly generated scenarios specifying a target object and its desired location, we employ a rule-based approach to create instructions that require spatial understanding. First, we extract the coordinates of all objects in the robot’s frame. We then generate spatial instructions by identifying the object(s) closest to the target object in a specific direction, ensuring that no other objects of the same type are present in the scene and that the angular deviation in other directions does not exceed 30 degrees. This method produces clear and unambiguous spatial instructions.

\noindent\textbf{Commonsense Instructions.}
We leverage GPT-4 to generate diverse instructions grounded in commonsense reasoning. For each task, we provide images from three perspectives (robot’s left view, right view, and hand view), along with details about the target object and its intended position. Instead of explicitly mentioning the target object, GPT-4 generates pick-and-place task instructions framed within a familiar daily scenario. We prompt GPT-4 to ensure that the scenario is uniquely tailored to the target object, reinforcing the need for commonsense understanding.

\section{Method}
\label{sec:method}

\subsection{Overview}  
Our motivation is to leverage grounding masks as an intermediate representation to enhance the generalization ability of robotic manipulation policies. This requires a large-scale, pre-trained vision-language model capable of generating grounded information about target objects and placement areas, referred to as the Grounded Vision-Language Model, discussed in Section~\ref{sec:method_glamm}. We then incorporate this grounding knowledge into the low-level policy network, where the grounded masks function as both an attention mechanism within the Grounded Perceiver and an additional input channel for vision data, as detailed in Section~\ref{sec:method_policy}. Finally, we outline the training and evaluation procedures for the complete framework in Section~\ref{sec:method_train}.

\subsection{Grounded Vision-Language Model}
\label{sec:method_glamm}
Recent advances in large vision-language models~\cite{glamm, lisa, florence-2, ferret-v2} have enabled generalizable, object-centric grounding. Our approach builds on GLaMM~\cite{glamm}, a state-of-the-art model that generates pixel-level segmentation masks, to guide the low-level policy network in accurately localizing target objects and placement areas.

\noindent\textbf{Base Model.}
The grounded vision-language model takes an image observation and a language instruction as input and outputs binary masks for target objects and/or target placement areas specified by the instruction, as shown in Figure~\ref{fig:model}(a). Given an image $x_\mathrm{v}$ and a text instruction $x_\mathrm{t}$, the image is encoded by the CLIP vision encoder~\cite{clip} to obtain the visual feature $\mathbf{F}_\mathrm{v}=\operatorname{CLIP}(x_\mathrm{v})$. This feature is then projected into the LLM's embedding space via an MLP projector $f_\mathrm{v}(\cdot)$, yielding $f_\mathrm{v}(\mathbf{F}_\mathrm{v})\in \mathbb{R}^{256\times D_\mathrm{t}}$. The LLM $\mathcal{L}(\cdot)$ integrates both the projected visual features and the text instruction to generate the output $y_\mathrm{t}$:
\begin{equation}
    y_\mathrm{t}=\mathcal{L}(f_\mathrm{v}(\operatorname{CLIP}(x_\mathrm{v})), x_\mathrm{t}).
\end{equation}
The model perceives the image through a prompt formatted as: ``{\fontfamily{qcr}\selectfont The $\texttt{<IMAGE>}$ provides an overview of the picture},'' where the $\texttt{<IMAGE>}$ token is replaced by the projected visual feature, represented as a sequence of 256 tokens.

\noindent\textbf{Pixel-level Grounding.}  
To achieve pixel-level grounding, we employ a fine-grained image encoder $\mathcal{E}(\cdot)$ and a pixel decoder $\mathcal{D}(\cdot)$ within the grounding head. The encoder is initialized with a pre-trained SAM~\cite{sam} encoder, while the decoder follows a SAM decoder-like architecture. A special token $\texttt{<SEG>}$ is introduced into the LLM's vocabulary to extract grounding-related features. The last-layer embeddings $\mathbf{F}_\mathrm{seg}$ corresponding to the $\texttt{<SEG>}$ token are projected into the decoder's feature space using a projector $f_\mathrm{s}$. The binary segmentation mask $M$ is then generated as follows:
\begin{equation}
    M=\mathcal{D}\left(f_\mathrm{s}(\mathbf{F}_\mathrm{seg}), \mathcal{E}(x_\mathrm{v})\right).
\end{equation}

For robotic manipulation tasks, the model is prompted to segment both the target object and the placement area (if applicable). As illustrated in Figure~\ref{fig:model}(a), the model’s output specifies these regions using separate $\texttt{<SEG>}$ tokens. Each $\texttt{<SEG>}$ token is decoded into an individual mask, yielding the target object mask $M_\mathrm{o}$ and the placement area mask $M_\mathrm{p}$, which are subsequently provided to the Grounded Policy Network.

\subsection{Grounded Policy Network}
\label{sec:method_policy}
After the grounded VLM translates the language instruction into masks for target objects and placement areas, these masks provide useful spatial guidance for the robot's policy. Rather than requiring explicit grounding of semantic descriptions, the policy can focus on leveraging this structured information to improve object localization and action execution. Given the strong generalization ability of the grounded VLM, the masks serve as a bridge between high-level language instructions and low-level manipulation strategies, helping the policy adapt to novel objects and environments more effectively.

\noindent\textbf{Base Model.}
For the policy network, we employ a language-conditioned transformer architecture, following the GR-1 model~\cite{gr1}. As shown in Figure~\ref{fig:model}(b), this model processes a sequence of historical image observations, robot states and a language instruction as input to predict future robot actions. Specifically, for each input image $x_\mathrm{v}$, along with corresponding masks $M_o$ and $M_p$, we apply channel concatenation and project the combined channels back to a size of 3 using a linear layer. The result is then fed into a pre-trained ViTMAE encoder~\cite{mae}. The encoded visual feature $\mathbf{Z}_\mathrm{v}\in\mathbb{R}^{197\times D_\mathrm{v}}$ is computed as follows:
\begin{equation}
    \mathbf{Z}_\mathrm{v}=\operatorname{ViTMAE}(\operatorname{Linear}(\operatorname{Concat}(x_\mathrm{v}, M_o, M_p))),
\end{equation}
where $D_\mathrm{v}$ denotes the hidden dimension of the vision encoder. The encoded feature $\mathbf{Z}_\mathrm{v}$ consists of a global representation $\mathbf{Z}_\mathrm{v}^\mathrm{CLS}\in\mathbb{R}^{1\times D_\mathrm{v}}$, obtained from the $\texttt{CLS}$ token, and a set of local patch representations $\mathbf{Z}_\mathrm{v}^\mathrm{P}\in\mathbb{R}^{196\times D_\mathrm{v}}$, corresponding to a $\texttt{14$\times$14}$ spatial grid. These patch features are then processed by the grounded perceiver, $\mathcal{P}(\cdot)$, which interacts with the grounding masks to reduce the number of tokens, ultimately producing the refined visual token features $\mathcal{P}(\mathbf{Z}_\mathrm{v}^\mathrm{P})$.

Additionally, the input language instruction $x_\mathrm{t}$ is encoded using the CLIP text encoder, yielding the text token feature $\mathbf{Z}_\mathrm{t}$. The input robot state $x_\mathrm{s}$ is projected through an MLP to obtain the state token feature $\mathbf{Z}_\mathrm{s}$. To enable action prediction, a learnable $\texttt{ACT}$ token with feature $\mathbf{Z}_\mathrm{a}$ is also included. Consequently, the token sequence for a single timestep input is structured as follows:
\begin{equation}
    \left(\mathbf{Z}_\mathrm{v}^\mathrm{CLS}, \mathcal{P}(\mathbf{Z}_\mathrm{v}^\mathrm{P}), \mathbf{Z}_\mathrm{t}, \mathbf{Z}_\mathrm{s}, \mathbf{Z}_\mathrm{a} \right).
\end{equation}
For a history length of $N$, the complete token sequence for one forward pass is constructed by aggregating token features from the past $N$ timesteps. This sequence is then processed by a transformer decoder, which predicts the next-step action tokens through the output $\texttt{<ACT>}$ tokens.

\noindent\textbf{Grounded Perceiver}  
A standard perceiver model~\cite{perceiver}, used in frameworks like GR-1, functions as a token resampler, reducing the number of tokens derived from the initial visual features. This is achieved through iterative attention layers between a small set of learnable query tokens and the original visual features. However, this token resampling process may lead to information loss, potentially limiting policy learning by failing to capture critical details about the target objects and placement areas. To address this, we propose guiding attention toward regions defined by grounded masks, ensuring that essential information is preserved for effective manipulation.

The perceiver takes as input the $\texttt{14$\times$14}$ patch features, denoted as $\mathbf{Z}_\mathrm{v}^\mathrm{P}$, extracted from the vision encoder, along with a set of learnable query tokens, $\mathbf{Q}_\mathrm{g} \in \mathbb{R}^{k \times D_p}$, where $k$ is the number of query tokens and $D_p$ is the token dimension. These initial queries, referred to as global queries, capture information from the entire image. To integrate grounded masks, we introduce two additional sets of query tokens: $\mathbf{Q}_\mathrm{o} \in \mathbb{R}^{k\times D_p}$ for the target object and $\mathbf{Q}_\mathrm{p} \in \mathbb{R}^{k\times D_p}$ for the target placement area. During each attention layer, these additional query tokens interact with the patch features, with attention guided by the respective masks, $M_\mathrm{o}$ and $M_\mathrm{p}$, applied via a mask fill operation (see supplementary material for details). This process produces the final set of perceived visual features, $\mathcal{P}(\mathbf{Z}_\mathrm{v}^\mathrm{P})$, with a total token length of $3 \times k$.

\subsection{Training and Inference}
\label{sec:method_train}

\noindent\textbf{VLM Fine-tuning.}
Although GLaMM was pre-trained on a large-scale grounding dataset, it may not precisely follow the prompt to identify the target object and placement area from a given manipulation instruction in a zero-shot setting. To address this, we construct an instruction-following dataset based on the generated simulation data to fine-tune the VLM. The grounded VLM model is fine-tuned using two types of losses: auto-regressive cross-entropy loss for text generation, and a linear combination of per-pixel binary cross-entropy loss and DICE loss for segmentation.

\noindent\textbf{Policy Training.}
In each forward pass, the policy network receives image observations, robot states over $N$ consecutive timesteps, and the corresponding language instruction. Based on this input, the network generates $N$ action tokens, which are processed through linear layers to predict arm and gripper actions. Since arm actions are continuous, we use Smooth-L1 loss $\mathcal{L}_\mathrm{arm}$ for optimization. For binary gripper actions, we apply Binary Cross Entropy (BCE) loss $\mathcal{L}_\mathrm{gripper}$. Thus, the total training loss for the policy network is: $\mathcal{L}_\mathrm{total} = \mathcal{L}_\mathrm{arm} + \mathcal{L}_\mathrm{gripper}$.

\noindent\textbf{Inference.} 
The inference process starts by prompting the grounded VLM to generate segmentation masks for the target objects and placement areas. These initial masks are provided to the grounded policy network to predict the next action tokens. To optimize inference time, segmentation masks are extracted from the grounded VLM only once, at the beginning of the episode. These masks are then used throughout the entire manipulation task as inputs to the policy network. This approach is well-suited for single-turn manipulation tasks, where the target object’s position remains stable until the gripper interacts with it, and the placement area remains fixed. Thus, the initial masks provide consistent guidance for the policy network throughout the task.

For multi-view image observations, each view is processed independently by the grounded VLM to obtain the corresponding masks, allowing for parallel processing. In the grounded policy network, each image and its corresponding masks are separately passed through the vision encoder and grounded perceiver to generate visual token features. During each forward pass of the policy network, only the action for the next timestep is predicted.

\section{Experiments}
\label{sec:exp}
\subsection{Simulation Setting}
Our simulation environment is built upon RoboCasa~\cite{robocasa}, a large-scale framework offering an automated scene generation pipeline. We leverage this pipeline to create diverse scene layouts and textures for data collection. Our main experiments are conducted within the RoboCasa environment, which includes a range of manipulation configurations. We classify the original RoboCasa data as ``Easy,'' due to its relatively simple scene complexity and straightforward instructions. To introduce more complexity, we generate challenging pick-and-place data that requires advanced semantic understanding and reasoning. The instructions are divided into three categories: ``Appearance,'' ``Spatial,'' and ``Common-sense,'' each representing different types of task-specific knowledge. Additionally, we include fundamental manipulation skills, such as opening and closing doors or drawers, pressing buttons, turning levers, and twisting knobs, to further challenge the model. In these tasks, target masks (e.g., for a drawer handle) are also generated to guide the robot’s policy in precise manipulation.

\begin{table*}[tbp]
\caption{\textbf{Performance Comparison on Simulated Tasks.} Metrics for pick-and-place tasks are reported as ``$a$ / $b$'', where $a$ is the contact rate (\%) and $b$ is the success rate (\%). We also report the success rates for other fundamental skills, including door opening/closing, button pressing, lever turning and knob twisting.}
\label{tab:main_results}
\resizebox{\linewidth}{!}{
\begin{tabular}{c|cccc|ccc}
\toprule
\multirow{2}{*}{Method} & \multicolumn{4}{c|}{Pick and Place} & \multirow{2}{*}{\makecell[c]{Open/Close}} & \multirow{2}{*}{\makecell[c]{Press}} & \multirow{2}{*}{\makecell[c]{Turn/Twist}} \\
 & Easy & Appearance & Spatial & Common-sense &  \\ \midrule
ACT~\cite{aloha} & 47.3 / 18.3 & 18.5 / 3.8 & 17.5 / 3.5 & 15.3 / 2.8 & 32.0 & 30.8 & 26.5 \\
BC-Transformer~\cite{robomimic} & 79.8 / 34.8 & 31.3 / 6.8 & 38.3 / 7.8 & 26.0 / 5.8 & 54.8 & 50.8 & 45.3\\
GR-1~\cite{gr1} & 85.3 / 42.8 & 49.5 / 13.8 & 54.5 / 16.3 & 43.0 / 11.5 & 66.8 & 58.8 & 49.5 \\
Ours & 89.0 / 43.3 & 78.5 / 30.5 & 81.0 / 33.5 & 76.3 / 30.0 & 72.0 & 69.3 & 54.5 \\ \bottomrule
\end{tabular}}
\end{table*}

\subsection{Main Results}
To evaluate our approach, we compare it with three well-established, easy-to-implement methods as baselines. Each method is trained on the same dataset, which includes RoboCasa’s 66K demonstrations for foundational skills, alongside our proposed dataset containing 24K demonstrations and 112K instructions with increased diversity and complexity. For evaluation, we generate 400 test samples for each instruction type within the pick-and-place task and for each fundamental skill. The results are presented in Table~\ref{tab:main_results}. Metrics for pick-and-place tasks are reported as ``$a$ / $b$'', where $a$ is the contact rate (\%) and $b$ is the success rate (\%). The contact rate refers to the percentage of attempts where the gripper makes contact with the target object, while the success rate indicates the percentage of tasks completed successfully. To ensure a fair comparison, we standardize key settings across methods, including history length, input image views, and image sizes.

\begin{table}[tbp]
\caption{\textbf{Evaluation Results of Unseen Settings.} Unseen instance denotes evaluation on new objects belonging to classes present in the training data. In contrast, unseen class refers to evaluation on objects from entirely new classes that were not included in the training data.}
\label{tab:unseen_results}
\resizebox{\linewidth}{!}{
\begin{tabular}{cc|cccc}
\toprule
Unseen Level & Mask & Easy & Appea. & Spatial & Comm. \\ \midrule
Instance & & 71.0 / 35.5 & 38.0 / 11.5 & 38.8 / 11.5 & 30.8 / 7.8  \\
Instance & \cmark & 85.8 / 42.5 & 75.5 / 29.5 & 77.8 / 31.8 & 72.0 / 28.0 \\
Class &  & 53.0 / 17.5 & 27.5 / 5.3 & 30.3 / 6.3 & 28.0 / 5.8 \\
Class & \cmark & 79.5 / 27.3 & 68.5 / 14.3 & 69.5 / 14.8 & 67.3 / 13.8 \\ \bottomrule
\end{tabular}}
\end{table}

\noindent\textbf{Baselines.}
\begin{itemize}[leftmargin=*]
\item \textit{ACT}~\cite{aloha}: A transformer-based policy network introduced by ALOHA~\cite{aloha}. Trained as a conditional VAE, it performs action chunking when predicting future actions.
\item \textit{BC-Transformer}~\cite{robomimic}: A standard behavior cloning policy built on a transformer architecture. We use the publicly available BC-Transformer implementation from RoboMimic, which integrates language input by encoding it with CLIP and combining it with vision representations through FiLM~\cite{film} layers.
\item \textit{GR-1}~\cite{gr1}: A GPT-style model designed for language-conditioned visual robotic manipulation. The original GR-1 model was pre-trained on a large video dataset and predicts both robot actions and future images. Since the pre-trained model is unavailable, we reproduce it here without large-scale pre-training or image prediction.
\end{itemize}

\noindent\textbf{Analysis.}
Compared to baseline models, our method consistently outperforms across all tasks. ACT generally performs poorly due to the absence of language input, which is crucial for task-specific language comprehension and reasoning. While BC-Transformer and GR-1 benefit from language input, their performance on more challenging tasks remains relatively low. This limitation likely arises from design shortcomings, as these models encode language input as a single, global text feature, which is inadequate for the nuanced understanding and reasoning required by semantically rich instructions.

Our approach, which leverages grounding masks to guide the policy, shows substantial improvements in challenging tasks, highlighting the effectiveness of incorporating such masks. Interestingly, we observe a consistent gap between the success rate and the contact rate, with the latter being significantly higher. This discrepancy suggests that the model's grasping capability could be further refined. We attribute this to the diverse range of objects in the dataset, which makes learning accurate grasping poses more difficult. Integrating pose-prediction networks, such as AnyGrasp~\cite{anygrasp}, could potentially help address this challenge. Nonetheless, the high contact rate demonstrates the effectiveness of our method.

\subsection{Zero-shot Evaluation.}  
For zero-shot evaluation, we categorize unseen settings into two levels: unseen instance and unseen class. Unseen instances refer to evaluation on new objects within classes present in the training data, while unseen classes involve evaluation on objects from entirely new classes not included in the training data. We exclude 30 classes, comprising 597 objects, as unseen and filter out 1/4 of the objects from each of the remaining classes, resulting in a total of 1,335 unseen instances. For each unseen setting, we generate 400 test samples for each reasoning type within the pick-and-place task. The evaluation results are presented in Table~\ref{tab:unseen_results}.

\noindent\textbf{Analysis.}  
The experimental results on both unseen instances and unseen classes show that incorporating mask information significantly enhances the model's generalization to novel settings, while maintaining strong performance on previously seen tasks. This improved generalization is reflected in better performance across new data distributions. Notably, in more challenging scenarios, mask guidance achieves approximately 100\% relative improvement over non-mask baselines, highlighting its crucial role in handling complex, unseen situations.

\subsection{Ablation Study.}
To accelerate our ablation study, we use a subset of the dataset, as full training and evaluation would require several days on 8 NVIDIA 4090 GPUs. This subset includes 3K original samples and 3K generated reasoning samples for a single pick-and-place scenario in RoboCasa. For evaluation, we generate 50 test samples per instruction type while maintaining the original diversity in scenes and objects to ensure representative and meaningful results.

\begin{table}[tbp]
\caption{\textbf{Ablation Study on Training Data and Grounded Masks.} ``Ori. Data'' refers to original data in RoboCasa, while ``New Data'' denotes our proposed data with diverse instructions. ``Pred'' indicates the use of predicted masks, whereas ``GT'' denotes ground-truth masks.}
\label{tab:ablation_data_mask}
\resizebox{\linewidth}{!}{
\begin{tabular}{ccc|cccc}
\toprule
Ori. Data & New Data & Mask & Easy & Appea. & Spatial & Comm. \\ \midrule
\cmark &  &  & 80 / 38 & 26 / 6 & 20 / 4 & 26 / 4 \\
 & \cmark &  & 78 / 34 & 42 / 12 & 52 / 16 & 40 / 12 \\
\cmark & \cmark &  & 90 / 60 & 46 / 16 & 54 / 20 & 38 / 12 \\
\cmark &  & Pred & 88 / 44 & 48 / 14 & 50 / 16 & 44 / 12 \\
 & \cmark & Pred & 86 / 42 & 72 / 32 & 78 / 38 & 70 / 32 \\
\cmark & \cmark & Pred & 90 / 64 & 70 / 40 & 80 / 38 & 74 / 36 \\
\cmark & \cmark & GT & 96 / 68 & 80 / 48 & 82 / 44 & 80 / 42 \\ \bottomrule
\end{tabular}}
\end{table}

\noindent\textbf{Training Data and Mask.}  
We perform ablation studies by training models with different datasets and mask configurations. The results, shown in Table \ref{tab:ablation_data_mask}, indicate that when trained solely on the original simple data, the model easily fits the ``Easy'' set—even without a mask—but struggles with more challenging sets. This supports our hypothesis that the limited complexity of the original data restricts the model's generalization ability. In contrast, training on our new dataset highlights the critical role of mask guidance in improving generalization. Furthermore, combining both original and new data enhances performance, suggesting that scaling the dataset could further strengthen grounding-aware policies.

\begin{table}[tbp]
\caption{\textbf{Ablation Study on Modules for Incorporating Grounding Masks.} ``Channel Concat.'' denotes whether to do the channel concatenation for the image and mask. ``Grounded Perceiver'' denotes whether to incorporate masks into the perceiver.}
\label{tab:ablation_mask_method}
\resizebox{\linewidth}{!}{
\begin{tabular}{cc|cccc}
\toprule
\makecell[c]{Channel\\Concat.} & \makecell[c]{Grounded\\Perceiver} & Easy & Appea. & Spatial & Comm. \\ \midrule
 &  & 68 / 24 & 42 / 12 & 52 / 16 & 40 / 12 \\
\cmark &  & 78 / 36 & 60 / 26 & 68 / 30 & 62 / 26 \\
 & \cmark & 80 / 36 & 56 / 22 & 64 / 28 & 64 / 30 \\
\cmark & \cmark & 86 / 42 & 72 / 32 & 78 / 38 & 70 / 32 \\ \bottomrule
\end{tabular}}
\end{table}

\noindent\textbf{Modules for Incorporating Masks.}  
We evaluate two methods for integrating grounding information: concatenating masks as an additional channel and using the grounded perceiver. As shown in Table \ref{tab:ablation_mask_method}, both approaches effectively incorporate grounding knowledge. While simple mask concatenation allows the model to utilize mask information, the grounded perceiver enables a more comprehensive exploitation of mask features, resulting in improved performance.

\begin{table}[tbp]
\caption{\textbf{Ablation Study on Grounding Representations.} ``Point'' and ``Bbox'' denote the center pixel point and the 2D bounding box extracted from predicted masks, respectively. ``Low-dim'' refers to using the point or bbox as a low-dimensional input, while ``Image'' indicates incorporating them as an additional channel in the image input.}  
\label{tab:ablation_representation}
\resizebox{\linewidth}{!}{
\begin{tabular}{c|cccc}
\toprule
Representation & Easy & Appea. & Spatial & Comm. \\ \midrule
Point, Low-dim & 72 / 28 & 44 / 12 & 56 / 18 & 42 / 14 \\
Point, Image & 76 / 32 & 60 / 26 & 68 / 32 & 64 / 28 \\
Bbox, Low-dim & 72 / 30 & 46 / 14 & 54 / 18 & 44 / 16 \\
Bbox, Image & 82 / 38 & 68 / 30 & 74 / 40 & 68 / 30 \\
Mask & 86 / 42 & 72 / 32 & 78 / 38 & 70 / 32 \\ \bottomrule
\end{tabular}}
\end{table}

\noindent\textbf{Grounding Representations.}  
We evaluate different grounding representations and their impact on model performance. As shown in Table \ref{tab:ablation_representation}, models perform best when using masks as grounding representations. In contrast, lower-dimensional vector representations, such as points and bounding boxes, pose greater challenges for effective learning. Masks provide richer information by accurately capturing the shape and size of the target object, leading to modest improvements over bounding box representations.

\begin{table}[tbp]
\caption{\textbf{Ablation Study on Grounded VLM.} ``Zero-shot'' refers to the zero-shot evaluation of the grounded VLM. ``Sim. data'' and ``VLM data'' denotes the use of simulated grounding data and VLM data for fine-tuning.}

\label{tab:ablation_glamm}
\centering
\resizebox{0.8\linewidth}{!}{
\begin{tabular}{l|c}
\toprule
Method & mIoU \\ \midrule
Zero-shot & 13.2 \\
Fine-tuned w/ Sim. data & 45.5 \\
Fine-tuned w/ Sim. data \& VLM data & 48.2 \\ \bottomrule
\end{tabular}}
\end{table}

\noindent\textbf{Grounded VLM}\label{sec:vlm_results}  
We fine-tune the VLM to improve its instruction-following capability by training it on both simulated and original VLM data. Specifically, we create an instruction-following dataset based on simulated data using the following prompt format: ``{\fontfamily{qcr}\selectfont Given a robotic manipulation instruction: \texttt{<Instruction>}, identify the target object for manipulation and, if applicable, the target placement area.}'' The response is formulated as ``{\fontfamily{qcr}\selectfont The target object is xxx $\texttt{<SEG>}$}'' to generate the target object mask, and ``{\fontfamily{qcr}\selectfont The target placement area is xxx $\texttt{<SEG>}$}'' if a placement area is specified.  

For evaluation, we compute the mean Intersection over Union (mIoU) of the predicted simulation results, with a performance comparison shown in Table~\ref{tab:ablation_glamm}. The initial zero-shot evaluation uses a dependency parsing tool to identify target nouns in the instructions, which are then used to prompt the VLM to generate a mask for each noun. However, this approach leads to suboptimal results due to ambiguities introduced by the dependency parsing. Fine-tuning on simulation data alone significantly improves results but risks losing the knowledge embedded in the original VLM dataset. To mitigate this, we fine-tune the model using both simulated data and the VLM's original grounding data, thereby enhancing simulation performance while preserving the model's prior knowledge.

\section*{Acknowledgment}
This work was supported by National Key R\&D Program of China (2022ZD0162000).

{
    \small
    \bibliographystyle{ieeenat_fullname}
    \bibliography{main}
}

\clearpage
\setcounter{page}{1}
\maketitlesupplementary

\renewcommand\thesection{\Alph{section}}
\setcounter{section}{0}

The supplementary materials are organized as follows:  
\begin{enumerate}[leftmargin=*]  
    \item Implementation details are provided in Appendix~\ref{appendix:implementation_detail}.  
    \item Data details are presented in Appendix~\ref{appendix:data_detail}.  
    \item A detailed explanation of the grounded perceiver is given in Appendix~\ref{appendix:grounded_perceiver}.  
    \item Discussions on limitations and future work are provided in Appendix~\ref{appendix:limitation}.  
\end{enumerate}

\section{Implementation Details~\label{appendix:implementation_detail}}
For the grounded VLM, we fine-tune GLaMM~\cite{glamm} starting from its publicly available checkpoint, pre-trained on the Grounding-anything Dataset, which contains 7.5M unique concepts spanning 810M regions. The fine-tuning process adopts an instruction-following approach to enable grounded conversational capabilities. To augment the model's training data, we incorporate 112K QA pairs generated from simulated data into its existing 277K grounded conversation dataset. The model is fine-tuned using LoRA with a rank of 8. The base learning rate is set to 3e-4 with a WarmupDecayLR scheduler, and the batch size is 20. The fine-tuning runs for 20 epochs, covering 10K training steps, and requires approximately 40 hours on 8 NVIDIA RTX 4090 GPUs.

For the grounded policy network, we re-implement the model architecture of GR-1~\cite{gr1}, omitting the image prediction head due to the unavailability of their video dataset for pre-training. Instead, we train the model from scratch using our simulation data. The training employs a base learning rate of 5e-4 with a cosine annealing schedule, a batch size of 32, and spans 5 epochs. Full training takes approximately 70 hours on 8 NVIDIA RTX 4090 GPUs. For faster experimentation during ablation studies, we use a subset of the data, reducing the training time to 5 hours and evaluation time to 1 hour.

\begin{figure*}[t]
\centering
\includegraphics[width=1.0\textwidth]{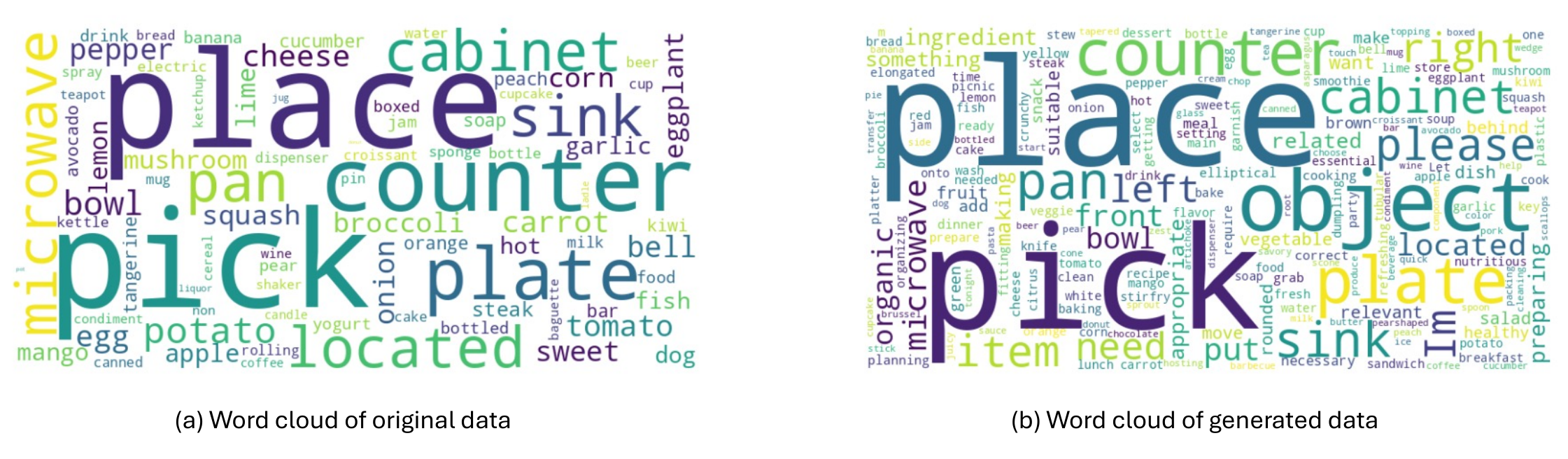}
\caption{\textbf{Comparison of Word Clouds: Original Data (left) vs. Generated Data (right).} }
\label{fig:wordcloud}
\end{figure*}

\begin{table}[pb]
\centering
\caption{\textbf{Embedding Similarity of Instructions: Original vs. Generated Data.} A lower mean similarity and higher variance suggest greater diversity in the generated data compared to the original data.}
\label{tab:inst_diversity}
\resizebox{0.8\linewidth}{!}{
\begin{tabular}{c|cc}
\toprule
 & Mean & Variance \\ \midrule
Original Data & 0.961 & 0.00043 \\
Generated Data & 0.888 & 0.00387 \\ \bottomrule
\end{tabular}}
\end{table}

\section{Data Details~\label{appendix:data_detail}}
\paragraph{Instruction Diversity.}
To assess the instruction diversity of our generated data, we first visualize word clouds for pick-and-place tasks in Figure~\ref{fig:wordcloud}, which clearly demonstrate the higher diversity of our generated data. Additionally, we quantitatively evaluate the diversity by using BERT to obtain CLS embeddings for each instruction. We then compute the cosine similarities for all instruction pairs and calculate the mean and variance of the similarity matrix as measures of diversity. The results, presented in Table~\ref{tab:inst_diversity}, show that the generated data exhibit lower mean similarity and higher variance, indicating greater diversity.

\begin{figure*}[t]
\centering
\includegraphics[width=0.8\textwidth]{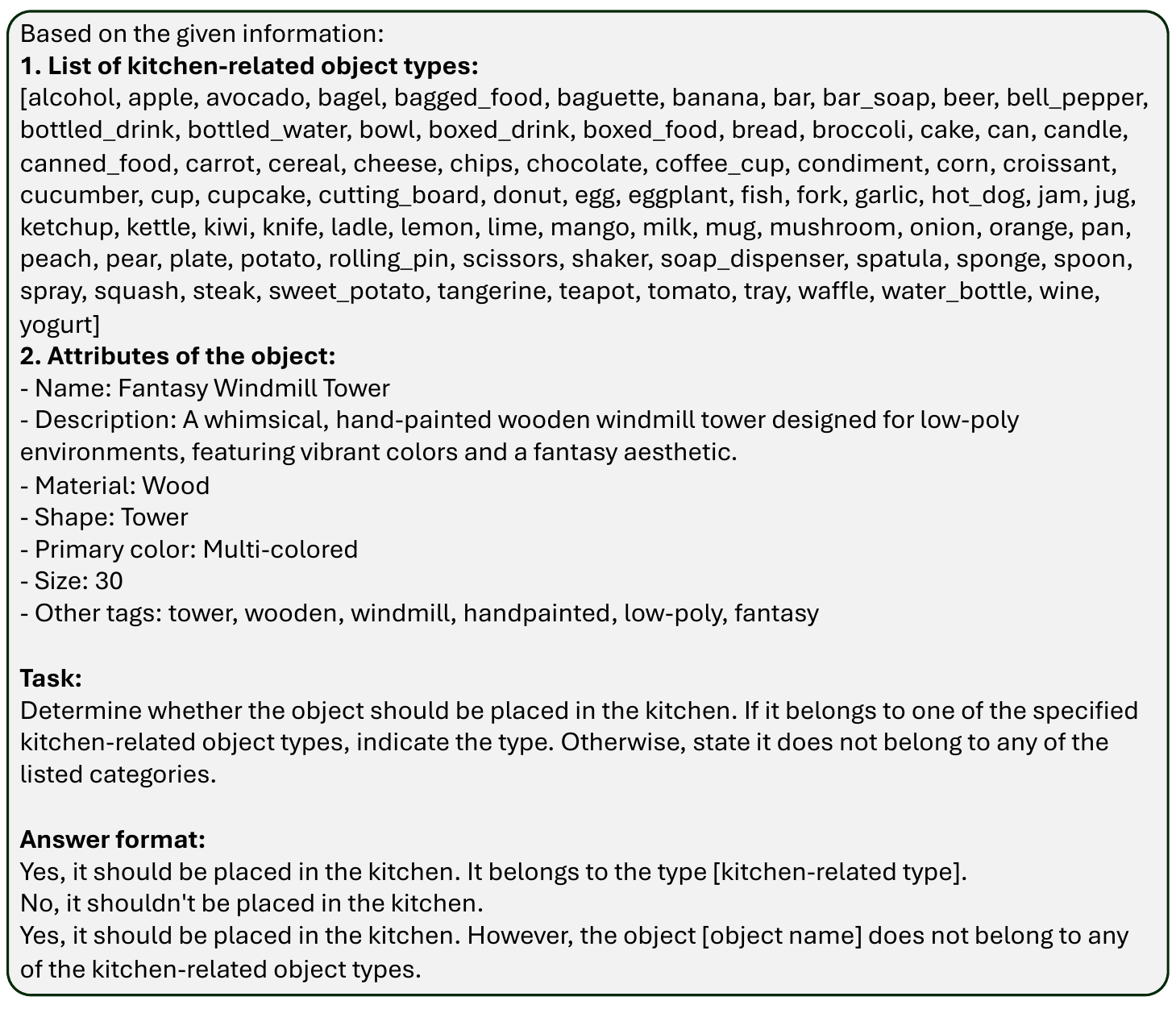}
\caption{\textbf{Prompt for Filtering Kitchen-related Objects.} }
\label{fig:prompt_obj_filter}
\end{figure*}

\begin{figure*}[t]
\centering
\includegraphics[width=0.8\textwidth]{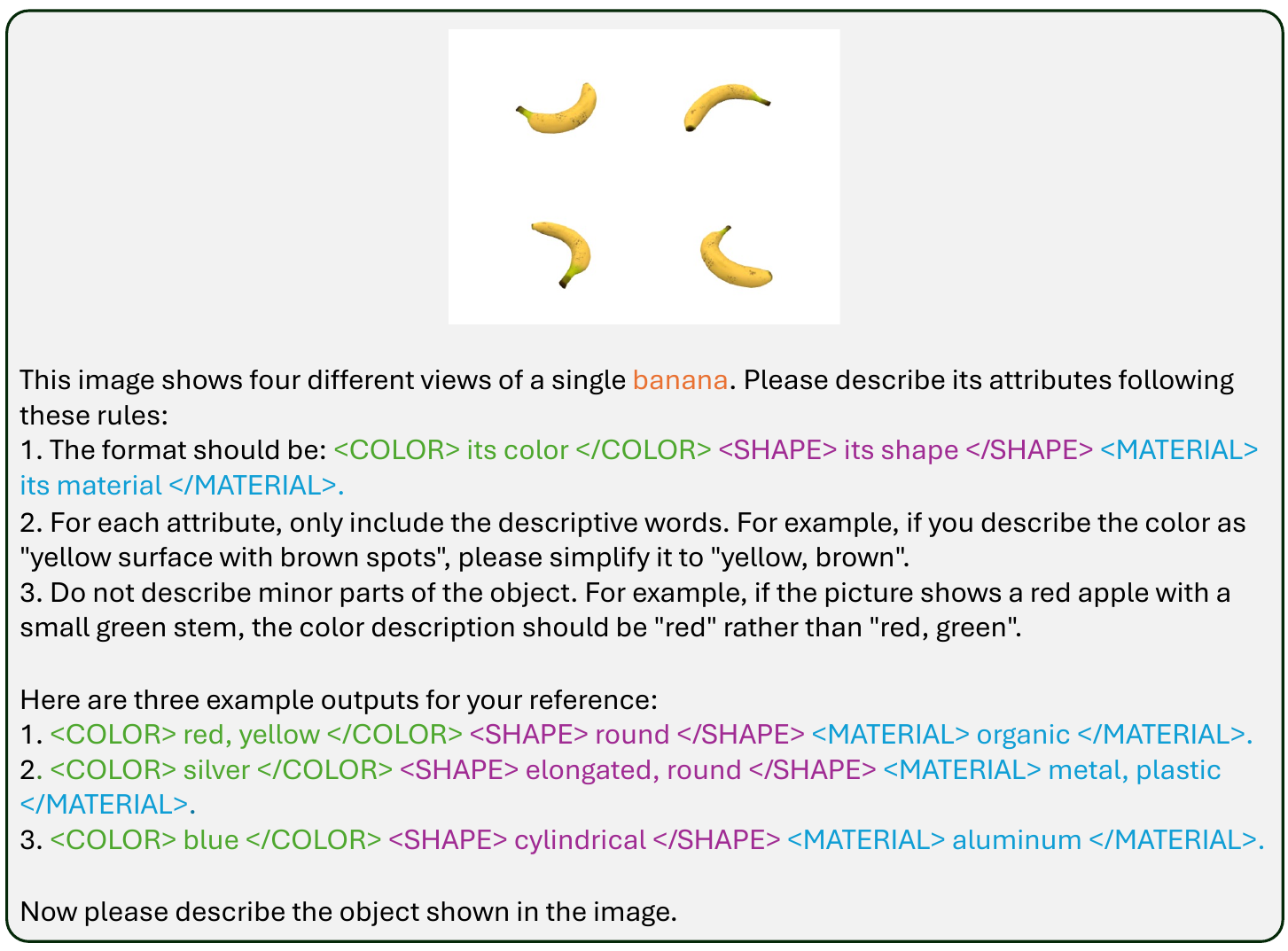}
\caption{\textbf{Prompt for Generating Key Attributes.}}
\label{fig:prompt_short_info}
\end{figure*}

\begin{figure*}[t]
\centering
\includegraphics[width=0.8\textwidth]{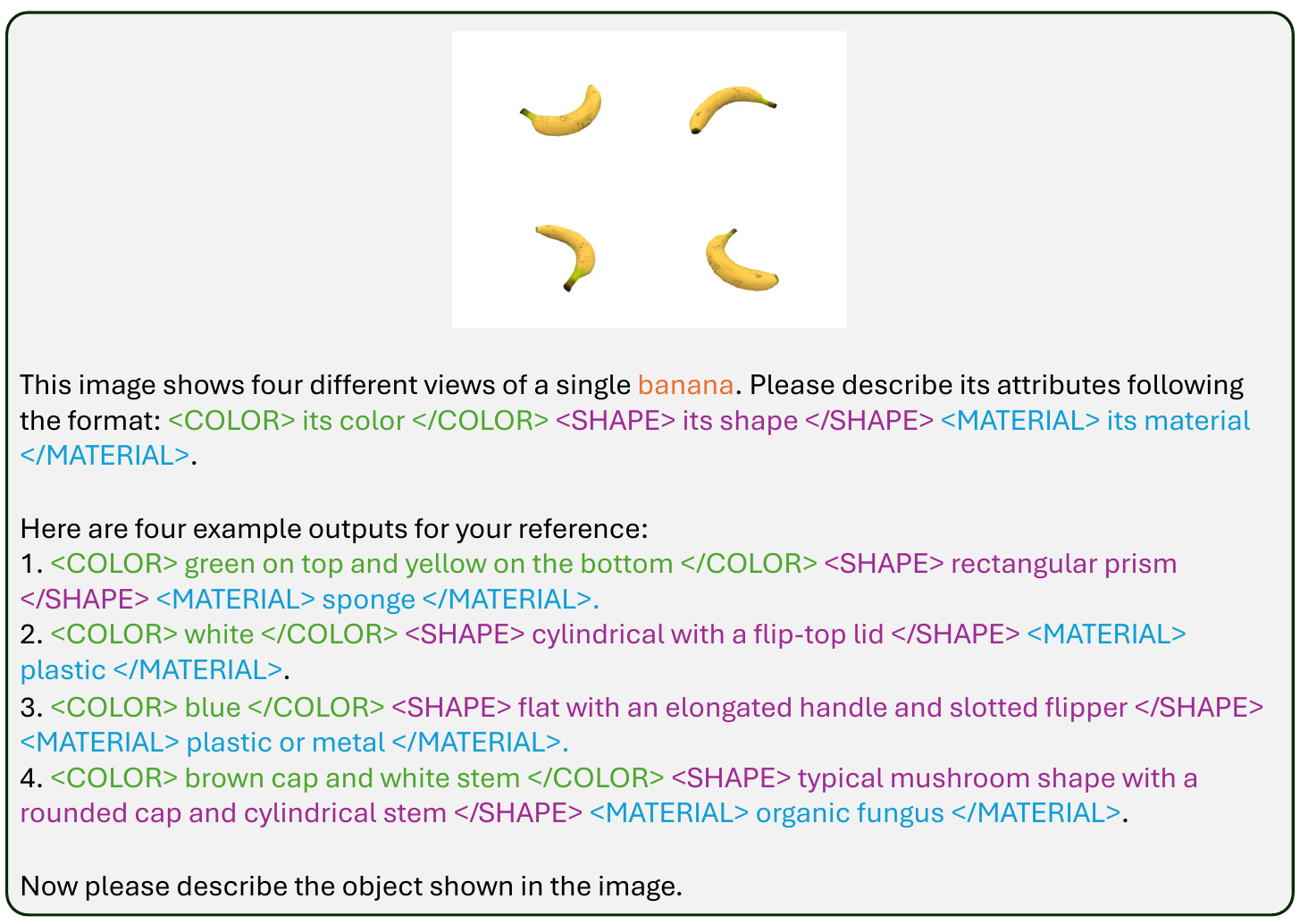}
\caption{\textbf{Prompt for Generating Descriptive Phrases.} }
\label{fig:prompt_long_info}
\end{figure*}

\begin{figure*}[t]
\centering
\includegraphics[width=0.8\textwidth]{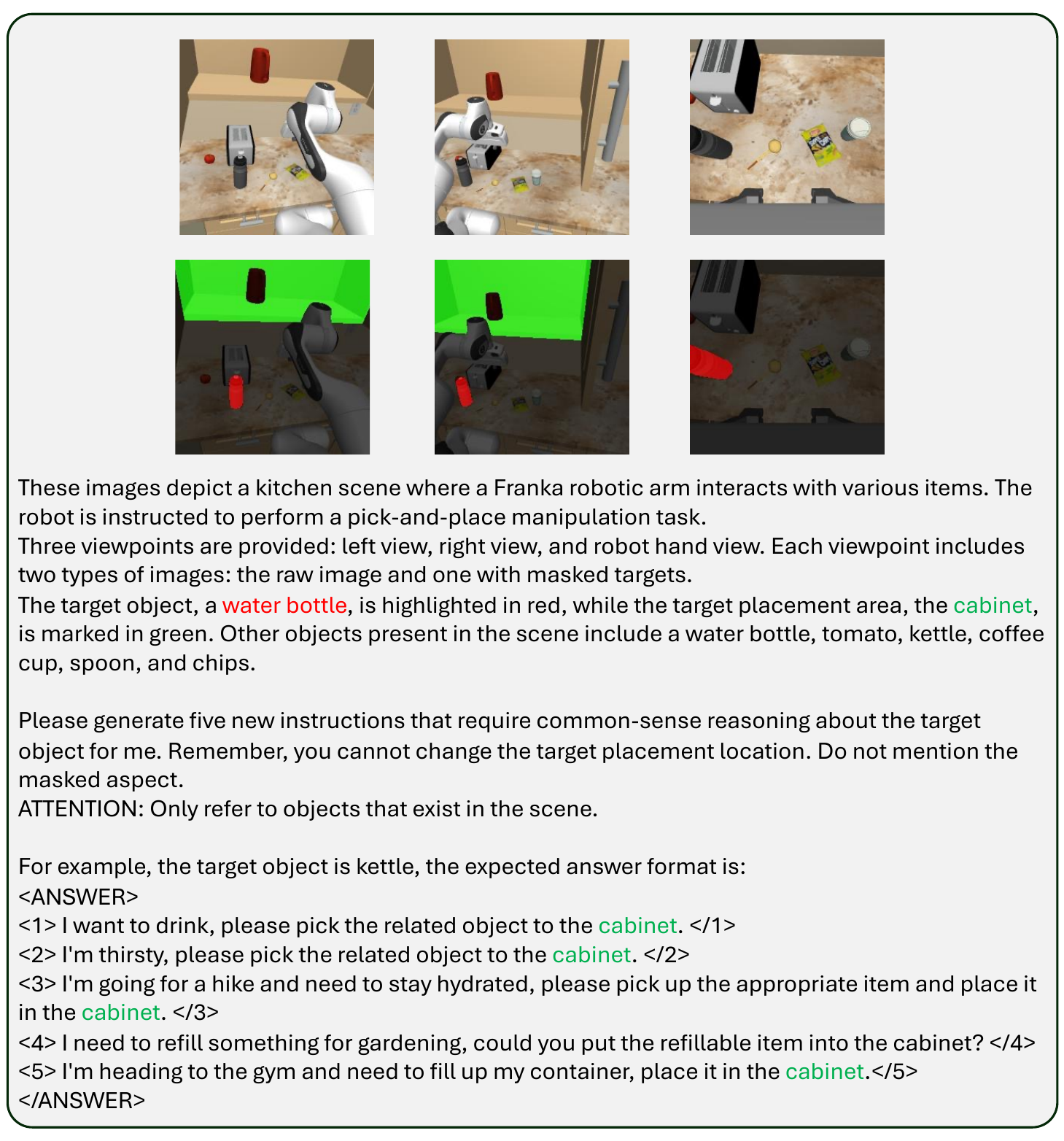}
\caption{\textbf{Prompt for Generating Common-sense Instructions.} }
\label{fig:prompt_common}
\end{figure*}

\paragraph{Prompts for Data Generation.}
We provide the prompts utilized for GPT-4 in our work. The prompt for filtering kitchen-related objects, shown in Figure~\ref{fig:prompt_obj_filter}, includes a list of valid kitchen-related object types and object attributes. GPT-4 is tasked with determining whether a given object is related to the kitchen based on this information. It is worth noting that the type list and object attributes were initially generated by GPT-4 in earlier stages of our process.

The prompt for generating key attributes of objects is illustrated in Figure~\ref{fig:prompt_short_info}. In this case, GPT-4 is instructed to describe the object’s attributes using distinct descriptive words, referred to as key attributes. Similarly, the prompt for generating descriptive phrases is depicted in Figure~\ref{fig:prompt_long_info}, where GPT-4 is tasked with describing object attributes using detailed descriptive phrases.

The prompt for generating common-sense instructions is presented in Figure~\ref{fig:prompt_common}. In this scenario, GPT-4 is provided with multiple views of the scene, including the target object, the placement area, and other surrounding objects. It is then instructed to generate instructions that require common-sense reasoning.

\paragraph{Simulation Tasks.}
We adopt the setup of 22 atomic tasks defined in RoboCasa~\cite{robocasa}, categorizing them into four task types:\textit{Pick and Place}, \textit{Open/Close}, \textit{Press}, and \textit{Turn/Twist}, as summarized in Table~\ref{tab:data_statics}. Beyond the original dataset of 3,000 generated samples for each task (referred to as ``Easy'' data), we introduce more complex scenes and instructions specifically for pick-and-place tasks. These enhancements aim to increase diversity by incorporating variations in appearance, spatial relationships, and common-sense reasoning.

\begin{table}[tbp]
\centering
\caption{\textbf{Task and Data Split.} We adopt the setup of 22 atomic tasks defined in RoboCasa, categorizing them into four task types. For pick-and-place tasks, we create more complex scenes and instructions to increase the level of diversity. The quantity of training data for each task type is detailed in the table.}
\label{tab:data_statics}
\resizebox{\linewidth}{!}{
\begin{tabular}{c|c|cccc}
\toprule
\multirow{2}{*}{Task Type} & \multirow{2}{*}{Task Name} & \multicolumn{4}{c}{Train Data} \\
 &  & Easy & Appea. & Spatial & Comm. \\ \midrule
\multirow{8}{*}{Pick and Place} & PnPCounterToCab & 3K & 3K & 5K & 6K \\
 & PnPCabToCounter & 3K & 3K & 5K & 6K \\
 & PnPCounterToSink & 3K & 3K & 5K & 6K \\
 & PnPSinkToCounter & 3K & 3K & 5K & 6K \\
 & PnPCounterToMicrowave & 3K & 3K & 5K & 6K \\
 & PnPMicrowaveToCounter & 3K & 3K & 5K & 6K \\
 & PnPCounterToStove & 3K & 3K & 5K & 6K \\
 & PnPStoveToCounter & 3K & 3K & 5K & 6K \\ \hline
\multirow{6}{*}{Open / Close} & OpenSingleDoor & 3K & - & - & - \\
 & CloseSingleDoor & 3K & - & - & - \\
 & OpenDoubleDoor & 3K & - & - & - \\
 & CloseDoubleDoor & 3K & - & - & - \\
 & OpenDrawer & 3K & - & - & - \\
 & CloseDrawer & 3K & - & - & - \\ \hline
\multirow{3}{*}{Press} & CoffeePressButton & 3K & - & - & - \\
 & TurnOnMicrowave & 3K & - & - & - \\
 & TurnOffMicrowave & 3K & - & - & - \\ \hline
\multirow{5}{*}{Turn / Twist} & TurnOnSinkFaucet & 3K & - & - & - \\
 & TurnOffSinkFaucet & 3K & - & - & - \\
 & TurnSinkSpout & 3K & - & - & - \\
 & TurnOnStove & 3K & - & - & - \\
 & TurnOffStove & 3K & - & - & - \\ \bottomrule
\end{tabular}}
\end{table}

\section{Details of Grounded Perceiver~\label{appendix:grounded_perceiver}}  
The grounded perceiver is composed of multiple attention layers. To illustrate its mechanism, consider a single attention layer where the input queries consist of 9 global query tokens $\mathbf{Q}_\mathrm{g} \in \mathbb{R}^{9\times D_\mathrm{p}}$, 9 target object query tokens $\mathbf{Q}_\mathrm{o} \in \mathbb{R}^{9\times D_\mathrm{p}}$, and 9 target placement query tokens $\mathbf{Q}_\mathrm{p} \in \mathbb{R}^{9\times D_\mathrm{p}}$. These queries are concatenated for parallel computation and projected to the hidden dimension of the attention layer, forming $\mathbf{Q} \in \mathbb{R}^{27\times d}$, where $d$ represents the hidden dimension. The input $\texttt{14$\times$14}$ patch features $\mathbf{Z}_\mathrm{v}^\mathrm{P} \in \mathbb{R}^{196\times D_\mathrm{v}}$ are concatenated with the query features to construct the Key $\mathbf{K} \in \mathbb{R}^{223\times d}$ and Value $\mathbf{V} \in \mathbb{R}^{223\times d}$. The attention matrix is computed as $\mathbf{A} = \frac{\mathbf{Q} \mathbf{K}^\top}{\sqrt{k}} \in \mathbb{R}^{27\times 223}$. To incorporate the masks for target objects and placement areas, the attention values corresponding to the masked regions are replaced with the highest attention value in the current matrix. Specifically, the target object masks $M_\mathrm{o}$ are applied to $\mathbf{A}_{[9:18, :196]}$, and the placement masks $M_\mathrm{p}$ are applied to $\mathbf{A}_{[18:, :196]}$. This ensures that the target object and placement query tokens focus more effectively on the relevant masked areas.

\section{Limitation and Future Work\label{appendix:limitation}}

Although extensive experiments have demonstrated the effectiveness of our proposed method using grounding masks as a guide—particularly its strong generalization ability to unseen domains—there remain some limitations that warrant further exploration in future work.

For object picking, we observe a significant gap between the contact rate and the success rate, suggesting that the model struggles with reliably grasping target objects. This limitation stems primarily from the high diversity of the thousands of objects in our dataset, making it more challenging for the model to overfit compared to previous, less varied datasets. Furthermore, while grounding masks excel at providing localization guidance, they offer limited support for enhancing grasping precision. To address this, a promising approach could involve integrating a pre-trained grasp pose prediction network, such as AnyGrasp~\cite{anygrasp}, which boasts strong generalization capabilities for novel objects. Incorporating such a network could enable the development of a more robust and generalizable policy network.

For data generation, our focus has primarily been on increasing the diversity of the target object, while largely overlooking the diversity of target placement areas. Enhancing this aspect can be achieved by generating a broader range of target placement options. To accomplish this, we need to collect additional human demonstrations for these newly generated scenes (trajectories to new placement areas) and leverage automated methods, such as MimicGen~\cite{mimicgen}, to further augment the dataset.

For model architecture, we currently treat it as two distinct components: a grounded VLM for mask prediction and a policy network for action prediction. Future exploration of end-to-end architectures or slow-fast systems could be both promising and challenging, potentially enabling more robust policies and harder tasks such as long-horizon tasks.


We hope our findings inspire further research into intermediate representations that can guide low-level policies, and provide valuable insights for generating more diverse scenes and instructions in robot manipulation.

\end{document}